# Exploring the consistency, quality and challenges in manual and automated coding of free-text diagnoses from hospital outpatient letters

Warren Del-Pinto[1], George Demetriou[1], Meghna Jani[2,3,4], Rikesh Patel[5], Leanne Gray[6], Alex Bulcock[4], Niels Peek[7,8], Andrew S. Kanter[9,10], William G Dixon[2,4,8], Goran Nenadic[1]

[1] Department of Computer Science, University of Manchester, UK
[2] Centre for Epidemiology Versus Arthritis, University of Manchester, UK
[3] NIHR Manchester Biomedical Research Centre, University of Manchester, UK
[4] The Northern Care Alliance NHS Foundation Trust, UK
[5] Manchester University NHS Foundation Trust, UK
[6] University Hospitals of Morecambe Bay NHS Foundation Trust, UK
[7] Division of Informatics, Imaging and Data Science, University of Manchester, UK
[8] Manchester Academic Health Science Centre, Manchester, UK
[9] Intelligent Medical Objects, Inc., Northbrook, IL, USA
[10] Department of Biomedical Informatics and Epidemiology, Columbia University, USA

## Abstract

*Objective*
To explore and evaluate the quality and consistency of manual and automated clinical coding of diagnoses from hospital outpatient letters.

*Materials and Methods*
Using a sample of 100 randomly selected outpatient clinic letters, two clinically-trained human coders performed manual coding of diagnosis lists to SNOMED CT. Automated coding of all diagnoses was performed using a commercial algorithm (IMO's Concept Tagger). A subset of 130 of the resulting annotated diagnoses that had been independently coded by both human coders were subject to a further evaluation with a panel of clinicians, with disagreements resolved through discussion, to decide upon a gold standard coding for each diagnosis. This gold standard dataset was used to evaluate the quality and consistency of coding performed by both human coders and computer software. Comparisons were made i) between the codes provided by each human coder, ii) between the human coders and the gold standard, and iii) between the computer software and the gold standard. An automated comparison was performed using a distance-based metric to quantify matches and to determine how many codes fell within certain distances of one another. A qualitative evaluation was then performed with the panel to decide whether each coding was "Good", "Acceptable" or "Not acceptable" in capturing the given diagnosis text. Correlation between the distance-based and qualitative metrics were also evaluated. Results were stratified according to whether the free-text diagnosis description contained one or multiple clinical findings.

*Results*
Independent coding by two human annotators led to exact matches in 73% of cases, increasing to 81% or 90% within a distance of one or two edges in SNOMED CT, respectively. Compared to the gold standard codes, human annotators had an exact match 78% of the time, with 86% within one position. This improved to 88% and 93%, respectively, when limited to text entries that included only a single clinical finding. The automated coding had an exact match of 61% compared to the gold standard, with 76% within one position. When limited to single diagnoses, this improved to 77% and 91% respectively. On average across the two human coders, 98% of codes were considered good or acceptable, as opposed to 88% of the computer-generated codes.

*Discussion*
Results have demonstrated that only three in every four free-text diagnoses were mapped to the exact same code by two independent annotators. In an equivalent task of comparing human and computer-generated codes to a gold standard code, humans slightly out-performed the computer, with both performing notably better when there was only a single diagnosis contained in the free-text description rather than multiple separate findings. Automated coding by the computer was considered acceptable in around nine in ten occasions.

*Conclusion*
Clinical coding is an inexact science, with full agreement between coders difficult to achieve even when provided codes capture the clinical intent to a high level. An automated process to convert free-text information about diagnoses to clinical codes performed nearly as well as humans and was considered acceptable 90% of the time.

# 1. Introduction

Healthcare is provided from a range of settings including general practice (GP), hospitals, pharmacies and care homes. Hospital outpatient departments provide specialist input to support the management of disease, advising patients and others in the healthcare system about diagnoses and management plans, often for long-term conditions. It is vital that documentation and communication are clear across different parts of the healthcare system, and that information is also conveyed clearly to patients. While some structured data is routinely collected, for example the occurrence of an outpatient visit in a particular speciality, further details of outpatient visits are often recorded solely as free text, such as in outpatient letters. The UK's Professional Records Standards Body (PRSB) has developed guidance about what should be included in outpatient letters and how they should be structured, including headings such as demographics, referrer details, diagnoses, medications, history, plan and requested actions. Figure 1 provides an example letter, where free-text provides communication about the patient's problems, including the clinician's thinking about the patient's diagnoses.

Dear Dr. O'Reilly

Thank you for referring Miss Gently to my rheumatology outpatient clinic.

| **Diagnoses** | 1. Multiple joint pain - no evidence of inflammatory arthritis, |
| | 2. Fatigue, |
| | 3. Sleep disturbance, |
| | 4. Type 1 diabetes, |
| | 5. Hypothyroidism. |

**History**
Miss Gently has had left wrist pain since December 2016. Since then she has also had right wrist pain and aching in the shoulders and knees. She describes tingling and burning in the forearms and in the calves and shins. Her symptoms are gradually worsening and they are now constant. She feels tired all the time and has broken, unrefreshing sleep. She has Type 1 diabetes and has been recently diagnosed with hypothyroidism – and has been put on thyroxine. Her inflammatory markers are normal.

**Figure 1: An example of an outpatient letter from the PRSB Outpatient Letter Standard.**
**(Professional Record Standards Body, 2018).**

Although the primary purpose of collecting healthcare information is to support direct care and communication between healthcare professionals, there are well established secondary uses of healthcare data. These include improving the quality of patient care by enabling better planning, audit and quality improvement projects (Neves, et al., 2019), as well as enabling research such as epidemiological studies (Williamson, et al., 2020). In some countries, clinical coding is also frequently used for billing purposes (American Academy of Professional Coders, 2022). Healthcare providers have therefore adopted the use of standardised clinical terminologies in many settings to support structured data capture (NHS Digital, 2020). This is common in some settings, for example GP surgeries, but is rare in others, such as outpatients, where free-text is often the only available source of data. However, free-text is not directly machine understandable, meaning that diagnoses and other information reported in outpatient letters cannot be viewed across a population of patients for any secondary use without additional processing. The lack of coded diagnosis data from hospital outpatient departments means, for example, that there is currently no national understanding of the distribution of diagnoses across patients in this setting, despite secondary care accounting for 72% of the annual NHS commissioning budget as of 2016 in the UK (Gainsbury, 2016). Consequently, outpatient-based services and thus long-term conditions have a significant challenge as their data capture does not easily support secondary use of real-world data, and therefore progress and understanding of such conditions may be hampered. To mitigate that, national audits of certain long-term conditions have been set up to fill this gap. While such audits provide important findings (Ledingham, et al., 2017), they often require bespoke data collection systems, duplicating data entry with significant additional time and resource, and also introducing possible transcription errors.

Mapping information within the text of an outpatient letter to a clinical terminology (often referred to as clinical coding) could provide the advantages of structured data capture, such as facilitating the interoperable storage, querying and exchange of clinical information among healthcare providers as well as population research. Depending on the provider and clinical setting, codes from different clinical terminologies may be used. Since 2018, the NHS has adopted the Systematized Nomenclature of Medicine Clinical Terms (SNOMED CT) terminology as a core terminology (NHS Digital, 2020), as required by the Health and Social Care Act 2012. It is used to capture clinically relevant information such as diagnoses, procedures, symptoms, family history, allergies, assessment tools, observations, devices and other content to code care delivery to individuals.

The process of manual coding, either by clinical teams in real-time or by dedicated clinical coding teams, requires substantial training (Varela, et al., 2022) and takes significant time as coding needs to be done for each individual patient and each encounter with the healthcare system. The transformation of large volumes of historical unstructured clinical documents, such as outpatient letters, into structured (coded) data presents an unfeasible manual challenge given the amount of unstructured information that exists within the healthcare domain and the available resources. Therefore, it is necessary to consider approaches to automatically map clinically relevant concepts from free-text to standardised clinical codes to alleviate this burden and unlock the potential of (decades of) information that is currently stored as unstructured data across the NHS by transforming it to structured, machine readable forms.

Text mining is a field of computer science that allows automated conversion of free-text information to structured, machine-understandable outputs (Savova, et al., 2010; Kraljevic, et al., 2021). The use of natural language processing (NLP) tools to extract and structure information from unstructured clinical text has been identified as one of the major areas of application for Artificial Intelligence in clinical care (Jiang, et al., 2017). Automated coding via NLP techniques has the potential to make the task of coding clinical documents practical at a large scale, which has led to extensive exploration of the topic recently (Gaudet-Blavignac, Foufi, Bjelogrlic, & Lovis, 2021; Ji, et al., 2022). The task is often formulated as multi-label classification on either document or mention level, as several codes might be assigned to a given piece of clinical text. However, while recent neural models have had remarkable success in many healthcare applications, the accuracy of automated clinical coding is still relatively modest, oscillating around 60% (Xie, Xiong, Yu, & Zhu, 2019; Dong, Suárez-Paniagua, Whiteley, & Wu, 2021; Dong, et al., 2022). Since clinical coding may not be perfect in all instances, it is important to understand how well both humans and text mining algorithms perform against a gold standard agreed upon by human clinicians. Currently there is not much work that evaluates the consistency of manual clinical coding, even within the same clinical settings (Dong, et al., 2022). In addition, existing gold-standard clinical datasets such as MIMIC-III (Johnson, et al., 2016) have been shown to be significantly under-coded (Searle, Ibrahim, & Dobson, 2020). The aim of this study was therefore to understand the comparability of manual and automated coding when converting free-text information about diagnoses to SNOMED CT codes using data from a dedicated diagnosis section within outpatient clinic letters, and to shed light to the coding differences observed between human coders and between the codes produced by automated software and those by human coders.

## 2. Methods and Data

In this paper we focus only on the semi-structured part of outpatient letters (see Figure 2), where a list of textual descriptions of diagnoses is provided. Each of these descriptions refers to one or more diagnoses and will be used as input for the clinical coding process. The main reason for using this list (as opposed to the narrative free text in the main body of the letter) is that we are interested in the quality and consistency of coding, both manual and automated, rather than in the assessment of both human and text mining capabilities to recognise and extract mentions of diagnoses in free-text narrative. Examining the quality of extracted diagnostic codes from the broader narrative in a letter, while important, lies outside the scope of this work. Nonetheless, insight gained from this work on the quality of coding of clinical diagnoses in text will also be informative for this broader task.

The coding task was to map each of these individual lines of diagnosis descriptions to one or more relevant SNOMED CT concepts that encode the clinical intent (see Figure 2 and the task specification below). For the purposes of this work, the focus was on diagnoses. As such, the mapping was restricted to only those concepts that fall under the *Clinical Finding* subhierarchy of SNOMED CT. For example, in Figure 2, entry 001 would be mapped to a single code for "*Seropositive Rheumatoid Arthritis*"; in entry 004, there are two diagnoses to be coded: "*Anxiety*" and "*Depression*", each with a separate SNOMED CT code. In this task, we did not consider procedures (e.g. "*Coronary artery bypass*") or test results (e.g. the value of estimated glomerular filtration rate (*eGFR*)), or any other clinical concept types.

Given a single line with free-text description, the result of the task is a set of SNOMED CT codes ("code set") that capture the clinical meaning, with respect to diagnoses, of the description. For example, given the free-text description "*Anxiety and depression*", the coder, either human or computer, may return the result [48694002, 35489007] which are the SNOMED CT codes corresponding to "*Anxiety (finding)*" and "*Depressive disorder (disorder)*" respectively. We refer to this as the *code set* provided by the coder for the given free-text diagnosis. A code set may contain one or several SNOMED CT codes.

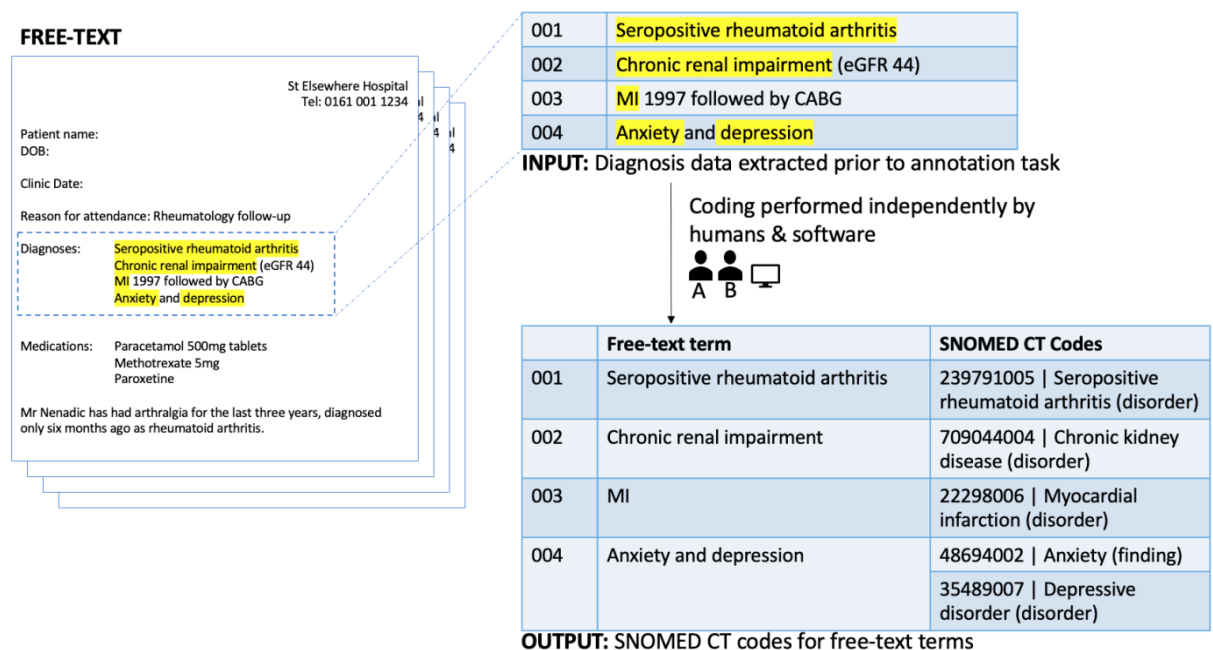

**Figure 2: An illustration of the conceptual pipeline for mapping diagnosis descriptions from free text to SNOMED CT codes.[1]**

The overall methodology of the work presented here consists of the three steps: 1. Data acquisition and preparation; 2. SNOMED CT clinical coding; and 3. Coding evaluation. These are explained below in detail. Ethical approval for the project was obtained by IRAS (212818) and REC (16/HRA/4393).

## 2.1 Data acquisition and preparation

A random sample of 100 outpatient letters from 2013-2017 was retrieved from the Rheumatology department of Salford Royal Hospital, which is part of the Northern Care Alliance (NCA) NHS Foundation Trust, one of the largest NHS trusts in the UK. From these, a semi-structured list of diagnoses was manually extracted from each letter by the Digital team at the hospital, and shuffled randomly within the list, so that subsequent free-text lines were unlikely to belong to the same patient. The diagnosis descriptions were checked manually so that they contained no sensitive or identifiable information. Any descriptions that included irrelevant content, such as formatting notes (e.g. "----") or empty lines, were excluded, resulting in a total of 708 free-text lines of diagnosis descriptions. The length of free-text diagnoses lines varied between 2 (e.g. "MI") and 188 characters, excluding spaces, with a median length of 28 and an interquartile range of 26 (a lower quartile of 18 and upper of 44). Additional descriptive statistics are available in Appendix (A.1).

## 2.2 SNOMED clinical coding

### The task and coding guidelines

The task was to code each line of textual diagnosis description to relevant SNOMED CT Clinical Findings, independently of other lines. With respect to the SNOMED CT ontology, we define *clinical findings* as any codes that are a descendent of "404684003 | Clinical Finding" in the SNOMED CT International Edition (as of Jan 2017). These codes include diseases (disorders), as well as other clinical findings such as symptoms (e.g. "*headache*"). The total number of such codes in the terminology was 107,465 as of Jan 2017. Although SNOMED CT includes the codes for other content to code care delivery to individuals, we decided to focus our investigations only on *clinical findings*, even if other information has been provided in free text and eventually coded by a coder. For example, given the following text:

"Previous right knee meniscal repair with secondary osteoarthritis"

the relevant information to be coded for this task was "*secondary osteoarthritis*", which could be mapped to the SNOMED CT code "443524000 | Secondary osteoarthritis (disorder)". However, the coding of other information including qualifiers, such as "*Previous*" and "*right*", and procedures such as "*meniscal repair*", was not necessary for the specified task. If these codes were included in the code set by a coder, then they were excluded from the analysis.

In collaboration with clinicians and coding experts, we developed a coding guideline (Appendix, A.5) that described the task and was used as a guide during the coding process. Specifically, the key coding principles included the following:

- The coding focus should be on clinical findings directly specified in a free-text diagnosis. No inference should be performed to derive or assume the existence of a diagnosis that is not directly mentioned or covered by the text. The only inference that should be performed during the coding process is the identification of an appropriate code for a disorder that is sufficiently described in the text, but uses different phrasing such as a synonym. For example, given *"scan showed Morton's Neuroma"*, a coder may choose to code this as "30085007 | Morton's metatarsalgia"[2] based on their own clinical judgement. If a free-text diagnosis mentions a clinical procedure or treatment, we do not infer the existence of a diagnosis unless the finding being diagnosed is explicitly stated in the wording of the procedure. For example, in "*cataract surgery*", "*Cataract*" is coded as a disorder, whereas in "*eye surgery*", there is no explicit mention of a disorder and therefore none should be coded.

- The chosen code should be as specific as possible so that it encodes the right clinical intent. If, however, the free-text description is unclear or underspecified, then a more generic code can be selected. For example, if a specific type of arthritis such as "*Psoriatic arthritis*" is not explicitly mentioned in the text, but might be (correctly or incorrectly) inferred by the clinician because of the presence of both "inflammatory arthritis" and "psoriasis", then the more general parent concept "*inflammatory arthritis*" should be used to annotate the text even though the more specific disorder could be inferred by the context.

- Pre-coordinated SNOMED CT terms should be used whenever possible. For example, "*severe asthma*" should be mapped to "370221004 | Severe asthma (disorder)", rather than mapping it to two separate concepts: "24484000 | Severe (severity modifier) (qualifier value)" and "195967001 | Asthma (disorder)". However, when coding a disorder for which there is not a single pre-coordinated code that fully captures the meaning of a clinical concept, the coder can add modifiers to the core concept, including locus/finding site, laterality, severity, chronicity/temporal associations, finding method and causative associations (such as coding "*infection*" in "*pancreatitis due to infection*").[3] However, the focus of the analysis is still on identifying and coding core disorder concepts, while additional modifiers are used for clarity during the coding process.

- When two or more distinct clinical concepts are present in the same narrative description, these should be coded as separate code sets (recall entry 004 of Figure 2, where "*Anxiety* and *Depression*" should be mapped to two separate SNOMED CT concepts).[4] We refer to these as "multi-finding" cases, as opposed to those that contain "single-finding" only.

---

[2] In the version of SNOMED CT used (2017), "Morton's metatarsalgia" is provided as a lexical synonym for "Morton's neuroma" under the same clinical code 30085007 | Morton's metatarsalgia (disorder).

[3] This is often referred to as post-coordination.

[4] Note that this is not post-coordination: cases with multiple codes are separate set of codes, rather than a single post-coordinated concept.

## Manual coding

For each of the textual descriptions in the dataset, the **manual** coding was performed by two clinically-active clinicians, referred here as "coder A" and "coder B". The coders were both practicing rheumatologists, with experience in using digital health technologies. The clinicians were asked to code each entry in the list of free-text diagnoses separately through the following steps:

- Identify core clinical findings in the free-text diagnosis, checking for multiple (distinct) clinical finding concepts appearing in a single free-text description.
- For each of these, use the SNOMED CT browser ([http://browser.ihtsdotools.org/](http://browser.ihtsdotools.org/)) to search for a pre-coordinated concept first, including trying synonyms, abbreviations or parent terms for the core concept. Once the core clinical findings are coded, the coder can post-coordinate the core concept with its qualifiers separately (although these qualifiers are excluded from the analyses we performed below).

Each coder performed the annotation task independently, providing a list of SNOMED CT codes for each line of free-text description. The list of 708 terms extracted from the letters was split in two subsets (for coders A and B). These two subsets had an overlap of 291 terms, which was coded by both coders. The coders were not aware of which terms were in the overlap. From this overlap set, a subset of 130 terms was used to create a gold standard (see below).

## Automated coding

For **automated** coding, we used IMO Concept Tagger, a commercial software solution developed by Intelligent Medical Objects (IMO), which specializes in developing, managing and licensing medical vocabularies. Specifically, IMO's clinical interface terminology maps diagnostic terminologies to SNOMED CT concepts by providing a bridge between terms used in clinical practice and standardized vocabularies. For each line of free-text diagnosis description, the text was sent to the software, which returned a coding of the text mapped to SNOMED CT. As with manual coding, the software only focused on the "Problem" codes, which correspond to SNOMED CT's "Clinical Finding".

## Gold standard

Following the code sets produced independently by each human coder, additional sessions to create an agreed gold standard were organised with a panel of four clinicians (the two original coders and two additional clinicians) and one independent assessor. The four clinicians were practicing rheumatologists, where the independent assessor was a general physician external to Salford Royal hospital. To perform the task, the panel had access to the SNOMED CT browser and to the codes provided by the coders A and B. Each of the free-text descriptions was discussed by the panel to produce a gold standard code set. If the panel members were not in agreement, the independent assessor adjudicated.

The gold standard was created from a subset of 130 clinical diagnosis text descriptions. Of these, 27 descriptions were judged to refer to concepts that were not clinical findings corresponding to a diagnosis (e.g. they may refer to procedures), or were too vague to be coded (e.g. "Allergy"). Since this work focuses on coding of diagnoses, these cases were excluded. An additional one description was found to be included twice in the dataset by mistake. Consequently, the gold standard set refers to the remaining 102 diagnoses.

## 2.3 Coding evaluation and metrics

Three sets of comparisons were performed over the code sets produced: i) a human-to-human comparison, ii) a human to gold standard comparison, and iii) a computer to gold standard comparison. The human-to-human case compares the code sets provided by coders A and B. The human-to-gold standard case compares the codes provided by coders A and B to the gold standard provided by the panel of clinicians, where the code sets provided by each individual coder for a given diagnosis description are separately compared to the corresponding gold standard. The software to gold standard case compares the codes generated by the IMO Concept Tagger (software) to the gold standard provided by the panel of clinicians.

In each of these comparisons, we performed two types of analysis: a distance-based evaluation and a qualitative analysis. When comparing between pairs of coders, we note that the number of examples compared may differ even when the comparison is made against the same dataset (e.g. the gold standard). This is because if one (or both) of the coders provided an invalid code set, for example one that did not contain any valid Clinical Finding codes, then the example would be excluded from the results. Therefore, depending on the pair of coders being compared, the final number of examples compared in the results may differ (see Table A.3 in the Appendix).

### Distance-based evaluation

We used the SNOMED CT hierarchy to evaluate the similarity between two code sets that have been provided for a given free-text diagnosis (e.g. one by a human coder and one provided as a gold standard). When each of the code sets contains a single code only, then the similarity can be calculated as the minimum distance (i.e. shortest path) between the two codes: the shorter the minimum distance, the more similar the two codes are. If any of the code sets being compared have more than a single code, then we must use a metric that calculates a distance between two *sets* of SNOMED CT codes (Girardi, et al., 2016). As we here consider diagnosis descriptions which may refer to multiple distinct clinical findings ("multi-findings" as mentioned above), the distance metric should only compare the corresponding codes in each code set that feasibly refer to the same finding in the diagnosis description, rather than "penalizing" the coding based on the diversity of findings described in the original text. To this end, we have defined the similarity as the average minimum distance between each code and its closest code in the other dataset. The full description of the distance-based evaluation can be found in Appendix (A.2). As an example, consider the following two code sets (derived for *"Left shoulder tendonitis and a fractured clavicle"*):

Set 1 = {202852009 | Shoulder tendinitis; 58150001 | Fracture of clavicle}
Set 2 = {76318008 | Disorder of tendon of shoulder region; 58150001 | Fracture of clavicle}

For the code "202852009 | Shoulder tendinitis", the closest corresponding code in Set 2 is "76318008 | Disorder of tendon of shoulder region". The distance between these codes is 2, which is calculated by following the path between the two codes, which is as follows:

> 76318008 | Disorder of tendon of shoulder region
> *parent_of*
>  239955008 | Tendinitis AND/OR tenosynovitis of the shoulder region
>  *parent_of*
>   202852009 | Shoulder tendinitis

where "*parent_of*" refers to a "type of" (subsumption) relationship in SNOMED CT. For the other code in Set 1 (58150001), there is an exact match in Set 2, resulting in a distance of 0. Therefore, the average minimum distance metric would return 1 when comparing these two code sets.

Given the complexity of evaluating code sets containing multiple Clinical Finding codes, we stratify the free-text descriptions into cases with only one Clinical Finding code ("single-finding" cases for which both coders provided a single *Clinical Finding* code to annotate the given text), and cases in which the code sets contain multiple clinical finding codes ( "multi-finding" cases). We also provide the results for "All" codes combined. Using the distance metric, we first analysed the number of exact matches, and then the number of matches that were within distance of 1, 2 or 3 or more of each other in the SNOMED CT hierarchy. We note again that we only focus on the *Clinical Finding* codes provided by each coder, and other codes are removed from the code sets.

## Qualitative analysis

Using our panel of clinicians, we also performed a qualitative analysis of the assigned codes. This investigated to what extent the codes assigned by a human coder or the computer algorithm represented clinical intent expressed by free-text description, according to the judgement of the panel, in particular when there was not an exact match to the gold standard. The following ratings were provided when assessing the overall quality of the code sets provided by each coder:

- Good: the clinical judgement is that the assigned codes captured the clinical findings appropriately (to a high level). For example, given the text "*Seronegative psoriatic pattern arthropathy (plantar fasciitis, Achilles tendonitis)*," an example of a "Good" coding is the following code set:

> 399112009 | Seronegative arthritis (disorder),
> 202882003 | Plantar fasciitis (disorder),
> 11654001 | Achilles tendinitis (disorder)

- Acceptable: the clinical judgement of the panel is that the code was relevant for capturing the main clinical intent for the diagnosis as a whole, although there might be some missing, broader/narrower, erroneous or inference-based codes in the set. For example, given the text *"Left side perineal tenosynovitis",* an example of an "Acceptable" coding is:

    67801009 | Tenosynovitis (disorder)

which is deemed to have a less specific (broader) meaning than the clinical notion described by the text.

- Not acceptable: the clinical judgement of the panel is that the code did not capture the clinical intent at an acceptable level. For example, given the text *"Mild colitis",* the following is an example of an "Not acceptable" coding is:

    128524007 | Disorder of colon (disorder)

as this concept is too broad to capture the specific meaning of the clinical notion in the text.

For each coder (human or software), we report the percentage of their codes that are considered "Good", "Acceptable", "Good" or "Acceptable", and "Not acceptable".

We also compared the results obtained using the distance metric to the qualitative analysis in order to determine the degree of correlation between the distance metric, based on the structure of the SNOMED CT ontology, and the qualitative evaluation provided by the expert panel. We first report the percentage of exact matches that were qualitatively categorised as "Good", "Acceptable" and "Not acceptable", and then we did the same for each of the distance of 1, 2 and 3 or more. Given that we are looking for a correlation between the distances and qualitative categories, we reported the results for all annotations together, rather than per individual coder.

## 3. Results

### 3.1. Distance-based evaluation

#### Human to human agreement

Table 1 shows the results of pairwise distance-based comparisons between the code sets provided by the human coders A and B (see also Table A.4 in Appendix). Nearly all of the code sets provided by coder A and coder B fell within an average minimum distance of 3 from one another. Significant portions (73%) of these were exact matches.

**Table 1: Distance-based comparison between the code sets provided by the human coders (A and B).**

|                | Exact match (%) | Distance (%) | | |
|----------------|-----------------|--------------|------|------|
|                |                 | <=1          | <=2  | <=3  |
| All            | 73              | 81           | 90   | 95   |
| Single-finding | 82              | 91           | 96   | 98   |
| Multi-finding  | 8               | 17           | 42   | 75   |

The results also indicate that consensus was more likely to be reached in annotating text when the clinical meaning could be captured by a single Clinical Finding code, and conversely notably less likely to be reached when the meaning required multiple Clinical Finding codes to capture.

## Human to gold standard agreement

The results in Table 2 shows the agreement between the two human coders and the gold standard (GS). The averages (Avg A/B) were obtained by micro-averaging.

**Table 2: Distance-based comparison between the code sets provided by human coders (A and B) and the gold standard (GS).**

|  |  | Exact match (%) | Distance (%) | | |
|---|---|---|---|---|---|
|  |  |  | <=1 | <=2 | <=3 |
| All | Coder A vs GS | 74 | 83 | 90 | 93 |
|  | Coder B vs GS | 82 | 89 | 95 | 96 |
|  | Avg A/B | 78 | 86 | 92 | 94 |
| Single-finding | Coder A vs GS | 86 | 92 | 96 | 98 |
|  | Coder B vs GS | 89 | 94 | 98 | 98 |
|  | Avg A/B | 88 | 93 | 97 | 98 |
| Multi-finding | Coder A vs GS | 0 | 23 | 46 | 62 |
|  | Coder B vs GS | 40 | 60 | 80 | 87 |
|  | Avg A/B | 21 | 43 | 64 | 75 |

The results show that in 78% of cases on average, the human coders matched the gold standard exactly, and that in 94% of cases the minimum average distance between corresponding codes in the human annotated set and the gold standard set was 3 or lower. Within the distance of 1 on single-finding cases, the agreement between the human coders and the gold standard was 93%. The correspondence between the clinicians' codes and the gold standard was significantly lower for the multi-finding cases: an average of 21% of code sets provided exactly matched the gold standard, while 75% fell within an average minimum distance of 3 or lower from the corresponding gold standard code set.

## Computer to gold standard agreement

Table 3 provides a distance-based comparison between the annotations produced by the software and the gold standard codes agreed upon by clinicians. The software gave an exact match in 62% of cases, 16 percentage points fewer than the clinicians. 12% fell outside the distance measure of 3 or more for the software, 6 percentage points more than for clinicians. However, within the distance of 1 in the single-finding cases, the agreement of the software with the gold standard (91%) was comparable to the same agreement level in case of the human coders (93%). Multi-finding cases performed less well, with 55% within a distance of three or less. Tables A.5 and A.6 in Appendix provide additional comparisons between the software and human coders.

**Table 3: Distance-based comparison between the code sets provided by the software and the gold standard.**

|  | Exact match (%) | Distance (%) | | |
|---|---|---|---|---|
|  |  | <=1 | <=2 | <=3 |
| All | 62 | 76 | 83 | 88 |
| Single-finding | 77 | 91 | 96 | 96 |
| Multi-finding | 0 | 15 | 30 | 55 |

## 3.2 Qualitative analyses

Table 4 provides the acceptability of codes provided by the human coders (A and B) and the software (Comp) as assessed by the clinical panel. On average, 85% of codes provided by the human coders fully capture the clinical intent ("Good"), with additional 12% providing an acceptable level. The acceptance of the software-generated code was around 10% lower for "Good" and for "Good" and "Acceptable" annotations.

**Table 4: Capturing clinical intent for the human coders (qualitative analysis).**

|  | Good % | Acceptable % | Cumulative (Good / Acceptable) % | Not acceptable % |
|---|---|---|---|---|
| Coder A | 83 | 15 | 98 | 2 |
| Coder B | 87 | 10 | 97 | 3 |
| Avg A/B | 85 | 12 | 98 | 2 |
| Comp | 75 | 14 | 88 | 12 |

## 3.3 Comparing the distance-based and qualitative metrics

Table 5 compares the results obtained using the distance metric, where the distance is taken from each code set to the corresponding gold standard code set, to the qualitative analysis. Since the aim is to compare the two approaches to assessing the acceptability of a given code set, rather than the performance of the coders themselves, the results are given across all examples and are not grouped by individual coders as for the previous results. In 86% of cases when clinicians provided a rating of "Good" for a given code set, the codes used were an exact match to the gold standard codes. An additional 10% of "Good" cases were a distance of 1-3 from the gold standard codes (see Examples 1 and 2 in Figure 3). There were no exact matches in the "Acceptable" and "Not acceptable" groups. Less than half of the "Acceptable" codes were within a distance of 1 in the SNOMED CT hierarchy, while one in three were more than three steps away despite being Acceptable. Of the small number of "Not acceptable" code sets, 60% fell within a distance of 3 from the gold standard, with some even being within a distance of 1 (see Example 3 in Figure 3).

**Table 5: Comparing the distance-based metric, from each provided code set to the corresponding gold standard, with the qualitative evaluation provided by the panel of clinicians.**

|  | Exact match (%) | Distance to Gold Standard (%) | | |
|---|---|---|---|---|
|  |  | <=1 | <=2 | <=3 |
| Good | 86 | 91 | 94 | 96 |
| Acceptable | 0 | 43 | 60 | 69 |
| Not Acceptable | 0 | 20 | 53 | 60 |

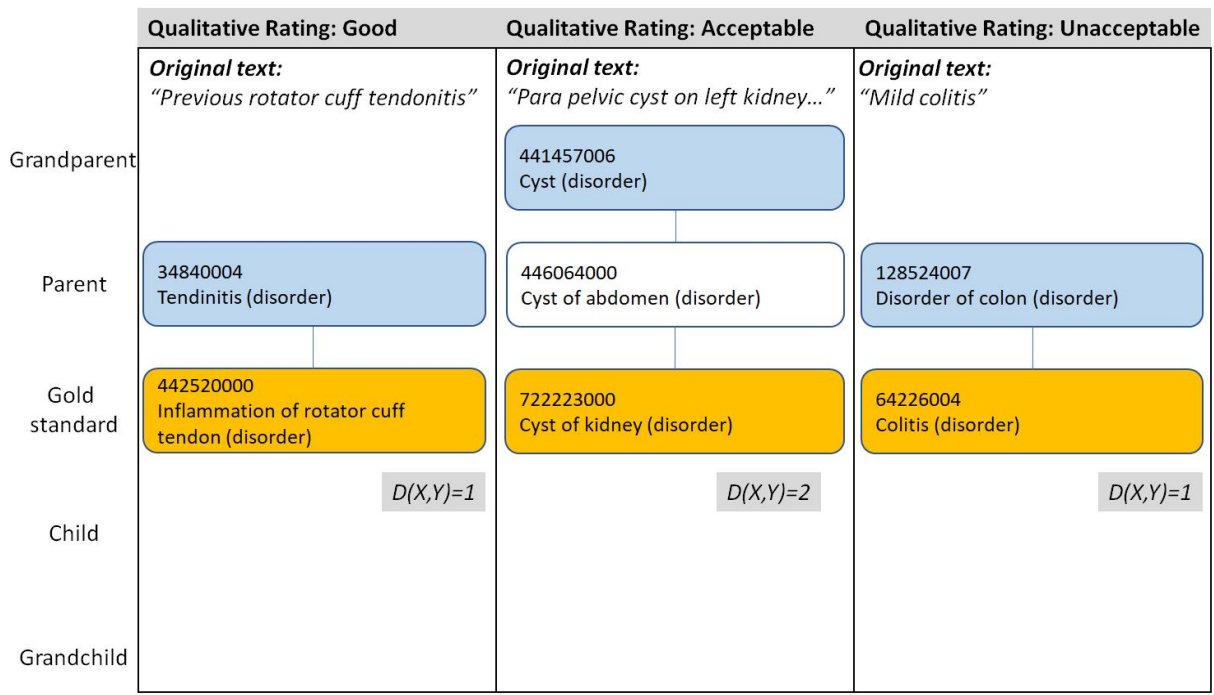

**Figure 3: Examples of the qualitative and distance-based assessments of code sets.** Example 1 (left): rated "Good" by the panel at a distance of 1 from the Gold Standard; Example 2 (middle): rated "Acceptable" by the panel at a distance of 2 from the Gold Standard; Example 3 (right): rated "Unacceptable" by the panel at a distance of 1 from the Gold Standard. Blue boxes indicate the codes provided by the coder, while yellow boxes represent the Gold Standard.

## 4. Discussions

Clinical coding is a challenging task for both human and automated coders. The aim of this work was to explore the similarities and differences in coding of given free-text clinical diagnoses to SNOMED CT Clinical Finding codes, as performed by different coders. Understanding how similar, or dissimilar, coding results are is important as this provides a way to assess the quality and consistency of both human-expert coding and computer-generated coding. Overall, the results indicate that clinicians agreed on exact codes for diagnoses contained in single line free-text descriptions in 3 out of 4 cases. Nearly all of the code sets provided by human coders fell within an average minimum distance of 3 from one another. These results indicate that human coders generally agreed on the clinical meaning of the presented text with respect to diagnoses, although they may have not always selected the same codes.

While in general the clinicians agreed or were close to the gold standard in the majority of cases, a small number of complex examples were notably more difficult to annotate consistently. Cases referring to only one clinical finding were notably easier to agree on than text containing multiple clinical findings. This may be due to the possibility that for the multi-finding cases, the clinical meaning was either less clear from the text, leading to difficulty in selecting a single code, or the meaning was more complex, necessitating the use of multiple codes to capture several distinct disorders, which has resulted in inconsistent choices. For example, a free-text entry "Seronegative psoriatic pattern arthropathy (plantar fasciitis, Achilles tendonitis, calcaneal oedema, good response to Kenalog)" requires identification of several clinical findings and may lead a coder to "ignore" some of the disorders mentioned. It is worth noting however that there were significantly more single-finding cases than multi-finding cases (7 times more), indicating that the majority of the presented text examples expressed a single clinical diagnosis with respect the type of *clinical finding* being referred to. Still, future guidelines for outpatient letters may be even more explicit in recommending the use of single clinical finding mentions, given that they are more suitable for communication and easier to clinically code.

When compared to a gold standard agreed between clinicians, the human coders provide an exact agreed code in 78%, with 94% of codes within the distance of 3 of the gold standard. The correspondence between the clinicians' codes and the gold standard was again notably lower for the multi-finding cases. We however note again that the overall number of multi-finding cases was relatively low, so these findings need to be taken with some caution.

The software-generated coding had fewer exact matches to the gold standard than human coders (62% as opposed to 78%), but the codes were within distance of 1 from the gold standard code in more than 90% of cases, and within the distance of 3 in over 96% cases for single-disorder descriptions. As for human coders, longer excerpts of free text that contained multiple diagnoses, qualifiers and post-coordinated terms generated more complexity, which affected coding quality and accuracy. While this stratification helped our understanding of the performance, we note that it does not allow us to selectively use automated coding on single-finding descriptions only as we will not be able to differentiate between a single- or multi-finding description before the automated coding.

While there were differences between coders, and while the coding by both human and computer coders was not always perfect, the qualitative evaluation provided by a panel of clinicians indicated that the codes still captured the key clinical intent of free-text diagnoses in the majority of cases: 98% of those generated by human coders and 88% of codes generated by computer were considered as "Good" or "Acceptable". Software-generated codes were rated as "Good" in 10% fewer cases than for individual human coders, with more unacceptable codes observed from the software, while the results were comparable between human and software for "Acceptable" cases. We acknowledge however that our categorisation is subjective: whether codes are "Good", "Acceptable" or "Not acceptable" in practice will depend on the specific use case for the coded data, and will differ by context (e.g. direct care versus informing different types of population health research). For instance, finding all patients with a specific subtype of rheumatoid arthritis for a departmental query would require more specific codes, as opposed to finding anyone with any type of arthritis for a research study, where an exact match is less important. In other words, the same imperfect match may be considered 'Good' for one use case, and 'Acceptable' or even 'Not acceptable' for a different use case. Our qualitative evaluation was not linked to a specific context or task, and therefore the estimates provided in this paper need to be interpreted with caution. It may be feasible, and indeed useful, to conduct future evaluations in the context of a particular use case.

When comparing the manual qualitative evaluation to the automated distance metric, we found that the distance metric was indicated to be a good proxy for the qualitative rating, particularly in the case of the exact match codes (all were considered 'Good'). Better qualitative ratings tended to correspond to code sets with a smaller distance from the gold standard codes, indicating that the use of automated distance-based metrics utilising the structure of the ontology could serve as a proxy to give an indication of the general quality of SNOMED CT codes assigned to free-text diagnosis descriptions. Nonetheless, small distances could still be considered "Not acceptable", or large distances "Acceptable". For example, example 3 in Figure 2 has a distance of 1 between the suggested and gold-standard codes, but was rated by the panel as "Not acceptable". On the other hand, example 2 in Figure 2 has a greater distance of 2 between the suggested and gold-standard codes, but has been rated as "Acceptable".

We have further examined whether the "easy" and "difficult" examples (for both the distance-based and qualitative metrics[5]) were the same for both human and software coders. In the case of distance-based comparison to the gold standard, in almost 60% of the examples for which a human coder provided the exact match to the gold standard, the software also provided the exact match. For almost all cases where the software provided an exact match to the gold standard, so did the human coders. On the other hand, in a small number of cases (around 6% on average), the software provided a coding that was more than three edges away ($D(X, Y) > 3$) from the gold standard, despite the fact that the human coder provided the exact match to the gold standard. This was the case mostly for multi-findings as the software either over- or under-coded. For example, in free-text description "STEMI (November 2012) - severe occlusion RAD stented, moderate LAD and moderate circumflex disease", the human coders have chosen the gold standard code "401303003 | Acute ST segment elevation myocardial infarction (disorder)"), while the software – in addition to that code – also added the additional codes for "Reactive airway disease" (991000119106) and "Coronary artery stenosis (233970002)", which has pushed the distance up (in part also because of the incorrect interpretation of RAD). This opens an interesting question as what should not be coded (e.g. because it is included in another code).

For the qualitative metric, the human coders (A and B) received the same qualitative rating from the panel in 85% (87/102) of cases. On average in 75% (76 / 102) cases, the coding sets provided by both the human coder and the software were of the same quality according to the panel's qualitative assessment. This means that in a quarter of cases, the level of complexity of free-text diagnoses was seen differently by the human and software coders. A

---

third of such cases where there was a mismatch between the ratings for the human and software coding sets were cases for which the human achieved a rating of "Good", while the software achieved a rating of "Not acceptable". Conversely, there was only one example (example 3 in Figure 3, "Mild colitis") for which a human coder received a rating of "Not acceptable" (as it was coded to "128524007 | Disorder of colon (disorder)"), while the software coding ("64226004 | Colitis (disorder)") was rated as "Good".

## Limitations

This work has several important limitations that need to be taken into account when interpreting the findings presented in this paper. Firstly, post-coordination was not considered for the coding of free-text diagnoses. As such, when mapping free-text to SNOMED CT codes, both human and software relied only upon pre-coordinated Clinical Finding codes. This is not a completely unrealistic setting in practice, as human coders in particular may want to focus on the most effective way to code by using pre-coordinated terms whenever possible. Still, this approach may present difficulty for the human coders who, depending on their experience with SNOMED CT, may rely upon the use of qualifiers to provide granularity to existing Clinical Finding codes. This is particularly true in the case where there is no suitable pre-coordinated Clinical Finding code available to capture the full meaning of the diagnoses occurring in a given text. Similarly, we note that the majority of automated clinical coding systems used to perform normalisation of free-text to SNOMED CT concepts do not include post-coordination. While systems may provide several codes for a given free-text description, they typically do not provide associations between disorder/finding terms and qualifiers. Post-coordination is however a strength of SNOMED CT: if coders have selected several codes for a given free-text description, it is possible to construct new expressions for meanings that are not captured by the terminology. Therefore, there is a need for annotation guidelines for both human coders and automated approaches, to capturing post-coordinated expressions from free-text, rather than relying on pre-coordinated codes only.

For this exercise, we created annotation guidelines (see Supplementary material) with instructions for coding to SNOMED-CT Clinical Findings alone. The SNOMED-CT vocabulary consists of various attributes including procedures, symptoms, morphological abnormalities, organisms, etc. Human coders sometimes mapped the given free-text diagnosis descriptions to SNOMED CT terms that matched closely to the original text, but did not include a Clinical Finding code, and these were considered as errors from the perspective of capturing diagnoses specifically. We also note that diagnosis descriptions were looked at in isolation (one at the time), without the complete clinical context, and different coders might have considered this lack of context differently, which makes the individual codes somewhat subjective. While we looked at diagnoses extracted from outpatient letters, we also note that the context under which the free-text was originally produced will also influence the coding process: for example, a clinician may convey additional information about the primary diagnosis in free-text format, either as a note to themselves or to help the recipient of the letter. Furthermore, the coding process is also influenced by 'local culture', i.e., longstanding ways of working that prioritise codifying certain information and the functionality of systems such as the availability of an EHR and how long it has been used in the given department.

The data used in this case study were from a single discipline (rheumatology) and from a single hospital. While the dataset contained both rheumatological and non-rheumatological diagnoses for patients being seen in that clinic, the data would inevitably have been weighted towards diagnoses in this discipline and it is an open question whether the findings would be generalisable to other settings.

All the coders were rheumatologists. This has potential implications for deciding on the 'gold standard', and perhaps even more so for what was considered Good/ Acceptable/ Not acceptable. There may have been a disease-specific bias when allocating acceptability: for example, "*shoulder tendonitis*" was coded by computer to "76318008 | Disorder of tendon of shoulder region" but that was deemed too broad. However, a clinician performing the same exercise from a more general perspective, or with a different medical speciality that focuses less on rheumatological diagnoses, may consider the same result as acceptable.

## 5. Conclusions

The use of standardised clinical terminologies such as SNOMED CT to represent patients' healthcare journeys is key for facilitating the primary and secondary use of clinical data: raw non-coded data (such as free-text narrative or images) are difficult to search and analyse. Manual coding of clinically relevant concepts represented in free text is not only time consuming, but also likely to be inconsistent. This difficulty is due in part to the importance of context in the medical setting: which information should be coded, and how the required information should be coded, depends on a multitude of factors including who the user or recipient of the coded data is and the purpose for which the data is to be used. For example, in clinical research settings, the type of information to be coded, and the level of granularity used for different subject areas, will likely differ from settings where data is used to directly inform patient care.

While there is a growing literature on clinical coding in general and to SNOMED CT in particular, there have been very few attempts to explore the consistency, quality and challenges in coding of real-world free-text diagnosis descriptions. The main purpose of this study was to better understand the challenges of clinical coding of diagnosis, and to shed light to the coding differences observed between human coders and between the codes selected by human and computer coders. The analyses of coding outcomes can be summarised by the following findings.

- Consistency of human annotations: clinical coders are not always aligned in selecting codes for a given free-text diagnosis: the inter-coder agreement between clinicians is 73% on exact matches, increasing to 81% if we look at codes that are within 1 node away from each other in the SNOMED CT hierarchy.
- Quality of coding: when compared to an agreed gold standard, the accuracy of human coders on average is 78% (exact matches) and 86% (within a distance <=1). The accuracy of computer-generated codes is 62% (exact matches) and 76% (within a distance of 1 to the gold standard).

- Capturing clinical intent: while there are inconsistencies and differences, still 98% of human and 88% of automated codes were considered as good or acceptable in capturing main clinical intent.

As clinical coding is a complex and challenging task for both human and automated coders, we note that there is a clear need to provide well-defined coding guidelines, as some inconsistencies in coding outcomes might be due to imprecise requirements or restrictions during the coding (e.g. the use of pre- or post-coordinated terms). This also includes the need to situate annotation guidelines for specific tasks under a common framework (Schulz, et al., 2023) as well as incorporating principles and developments in representation formalisms for structuring clinical knowledge (Rector, Qamar, & Marley, 2009; Bodenreider, Cornet, & Vreeman, 2018; Schulz, Stegwee, & Chronaki, Standards in Healthcare Data, 2019; Ayaz, Pasha, Alzahrani, Budiarto, & Stiawan, 2021). These considerations ensure that the results of manual or automated coding of clinical free-text are interoperable within healthcare systems and can be meaningfully compared to other annotated clinical text datasets.

Further work is also required to explore how coding can be made more efficient in practice, including post-hoc coding (coding of existing diagnosis descriptions). Our findings demonstrate that textual descriptions that refer to a single disorder are notably easier to code by both human and automated coders, so future recommendations may suggest that outpatient letters should strive to have one diagnosis per line to facilitate both more efficient automated future coding.

Our analyses also demonstrated that the current clinical coding practices could provide outcomes that – although not ideal – could be used to support a number of tasks, including epidemiological research that could benefit from computer coding on a large scale. Still, we note that the evaluation of the quality and acceptability of coding needs to be placed in a specific context and scenario of a use case, rather than being considered irrespective of what the codes will be used for. We also acknowledge that such evaluations are naturally subjective and that further work is required to develop task-specific evaluation settings. The ability to provide clinical coding on a large scale (e.g. by using semi- or fully-automated software-generated codes) will transform future population health research that will has access to coded diagnoses from different hospital specialists, as opposed to access to only coded data from GPs. Nonetheless, we still need to explore to what extent such codes will be useful for research, or indeed in clinical care, and what challenges this approach would bring to specific settings, and how these need to be mitigated.

## Acknowledgements

This work has been initially supported by the ISACC (IMO-Salford Automation of Clinical Coding) project, funded by Intelligent Medical Objects (IMO), followed by funding by the Farr Institute for Healthcare Research, the UK Healthcare Text Analytics Network (Healtex, funded by EPSRC EP/N027280/1), "Assembling the data jigsaw: powering robust population research in MSK disease" (funded by Nuffield Foundation) and "Integrating hospital outpatient letters into the healthcare data space" (funded by UKRI/EPSRC, grant EP/V047949/1). We are grateful to Prof Iain Buchan for initiating the project; Dr Sabine Van Der Veer (University of Manchester) for leading the ethics application for the project; Dr Azad Dehghan for setting-up the project; the NCA/SRFT Data science/BI teams for extracting data and installing the software; SNOMED International for providing initial training and discussions; the IMO team for providing technical support; the legal and project management teams at the University of Manchester, NCA and IMO for sorting out the agreements. The project has been part-funded by the Nuffield Foundation (visit www.nuffieldfoundation.org), but the views expressed are those of the authors and not necessarily the Foundation. MJ is funded by a National Institute for Health Research (NIHR) Advanced Fellowship [NIHR301413]. The views expressed in this publication are those of the authors and not necessarily those of the NIHR, NHS or the UK Department of Health and Social Care.

## Data availability

Due to ethics restrictions, access to the data used in this study needs to be requested from the Salford Royal Hospital and the Northern Care Alliance NHS Foundation Trust. The authors can assist in that process, but cannot provide the data themselves.

# Appendix

## A.1: Dataset descriptive statistics

Table A.1 contains the description of the dataset used in this paper.

| | |
|---|---|
| Number of free-text diagnosis descriptions | 708 |
| Mean length (characters) | 36 |
| Min length of description (characters) | 2 |
| Max length of description (characters) | 188 |
| St. dev of length (characters) | 27 |

Table A.1: Statistics for 708 free-text diagnoses extracted from 100 outpatient letters.

Table A.2 shows the number of examples that have been coded manually by each of the coders (A and B). A subset of 291 examples were coded by both coders (independently). The entire set of 708 examples was coded by the software.

| | Number of diagnosis coded |
|---|---|
| Coded by both A and B | 291 |
| Coded by coder A only | 191 |
| Coded by coder B only | 226 |
| Total terms manually coded | 708 |

Table A.2: Statistics for the manual annotation

## A.2: Distance-based comparison between coders

To evaluate the similarity between two code sets that have been provided for a given free-text diagnosis, it is necessary to define a suitable notion of similarity. Given the resource intensive nature of manual evaluation of code sets, it is beneficial to make use of a metric that can be automatically computed over the terminology from which the codes are derived. Additionally, since this work does not evaluate coding with respect to a specific application, it is necessary to define a generic metric that provides an indication of how similar two code sets are in general.

There are several important factors that should be taken into account by the metrics that are used to evaluate the similarity between the codes provided by each coder. In particular, the chosen metric:

- Should be able to account for sets of codes, rather than just single codes.
- Should not penalise based upon the number of codes used to annotate the given text, i.e., the evaluation should be consistent across sets of codes of varying sizes.
- Should not penalise based upon the diversity of codes used to annotate text, provided that there are similar codes across both of the sets being compared. It may be the case that a range of different diagnoses exist within the text and therefore similarity between two annotations should be based upon the presence of similar codes pointing to each separate diagnosis. Effectively, the metric should consider "subgroupings" of codes when measuring similarity.

- Should take into account "uncovered" codes, i.e., if one coder has identified a diagnosis and represented this by a code that is completely unrepresented in the set provided by the other coder.

Taking these points into consideration, the distance metric used for this evaluation is as follows:

$$D(X,Y) = \frac{1}{|X \cup Y|} \left( \sum_{x \in X} min_{y \in Y} d(x,y) + \sum_{y \in Y} min_{x \in X} d(y,x) \right)$$

where X, Y are the sets of codes provided by each coder and *d(x, y)* denotes the minimum distance (shortest path) between codes *x* and *y*. Note that, given two code sets, for each individual code only the closest code in the other set is considered as part of the calculation. Effectively, the distance metric can be interpreted as the *average minimum distance* between each code and the closest code in the other code set.

Effectively, the metric attempts to group the codes in each set by the clinical concept they represent, calculating similarity based upon the closest match (minimum distances) rather than simply the distance to every other code in the other set.

It is worth noting that, if the metric D(X, Y) returns a value of 0, then this corresponds to the case in which the two sets contain exactly the same codes, i.e., the annotations provided by the two coders match exactly.

To perform the pre-processing and analysis, the January 2017 release of SNOMED CT International Edition was required in Web Ontology Language (OWL) format. The release format 2 (RF2) files for this release were converted to OWL format using the tool available at https://github.com/IHTSDO/RF2-to-OWL

This conversion was necessary to make use of the OWL API, and the description logic reasoner ELK, to reason over the SNOMED CT ontology. The implementation used was written in Java.

The implementation was used for the following:

- **Restricting the data to only the relevant clinical codes**. This was done by checking whether or not a given code was a descendent (subclass) of the *404684003 | Clinical finding (finding)* SNOMED CT concept using ELK. If a code was not a subclass of Clinical Finding, it was excluded during pre-processing. If a given annotation result did not contain any *Clinical Finding* codes, then the result was excluded from the analysis.
- **Calculating distance metrics for the comparisons of annotation results**. These distances were calculated by classifying the SNOMED CT ontology using ELK. Effectively, this provides a graph in which each node is a concept (code) and each edge is a subsumption ("is-A", subclass) relationship between two codes. Distances between two codes can then be calculated by traversing the graph (calculating the distance to and from the least common ancestor of two codes). Over this graph, the *minimum distance* between two codes can then be defined as the shortest path between their corresponding nodes. This can be extended to code sets as given by the distance metric described earlier. Note that the graph includes implicit (as well as explicit) relationships between codes, due to the use of the OWL reasoner.

As an example of the pre-processing step, given the text *"Osteoarthritis - multi level degenerative changes"* the coder might provide the code set containing the codes 396275006 | Osteoarthritis (disorder)[6] and 33359002 | Degeneration (morphologic abnormality). Since the second code is not a *Clinical Finding* code, the code 3335902 is removed from the original result, where the resulting code set is then taken as the final coding for this description and coder for the purposes of the analysis. Similarly, for code sets containing codes that were erroneous, i.e., absent from the terminology (perhaps due to typographic errors while coding), then these were also removed during pre-processing.

The notion of distance is still applicable when comparing code sets of different sizes. For example, given the text "Previous right knee meniscal repair with secondary osteoarthritis" and the following code sets:

Set 1 = {239873007 | Osteoarthritis of knee; 443524000 | Secondary osteoarthritis}
Set 2 = {239873007 | Osteoarthritis of knee}

The first code in code set 1 will be compared to the closest corresponding code in set 2, which is in this case an exact match (distance 0). The second code, "443524000 | Secondary osteoarthritis" will also be compared to the closest corresponding code in set 2, which in this case means it will also be compared to "239873007 | Osteoarthritis of knee". The path between these two codes is via a common ancestor, "396275006 | Osteoarthritis" as follows:

      443524000 | Secondary osteoarthritis
       is_a
       396275006 | Osteoarthritis

and

      239873007 | Osteoarthritis of knee
       is_a
       396275006 | Osteoarthritis

resulting in a distance of 2. Therefore, the average minimum distance between the two code sets is 1. As such, the metric indirectly penalises the absence of "Secondary osteoarthritis" in the second code set. In some cases, this behaviour results in a larger distance (and hence "less ideal") between two code sets when there is missing information in one set. However, there may be cases where a difference between the size of two code sets does not necessarily imply missing information: multiple codes may be used to express the same or similar information to a single code. Additionally, the importance of missing information depends upon what the information is and the application for which the code sets are being evaluated.

---

[6] Note that *disorder* codes fall beneath the *Clinical Finding* hierarchy of SNOMED CT.

## A.3 Distance-based comparisons - additional results

When pairwise comparing a number of code-sets between two coders (two human coders, a human and software coder, or a coder and the gold standard), we note that the number of examples differed depending on which pair of coders was being compared. The reason for this is that a comparison could not be made if one or both of the coders did not provide Clinical Finding codes for a given example. The number of examples where the both coders have provided Clinical Finding codes for the same free-text description is used as a denominator in the corresponding calculations. Table A.3 shows the numbers of examples used for specific pairwise comparisons.

| Pairwise Comparison | Total examples compared | Single-finding examples | Multi-finding examples |
|---|---|---|---|
| A vs B | 97 | 85 | 12 |
| A vs GS | 98 | 85 | 13 |
| B vs GS | 99 | 84 | 15 |
| Comp vs GS | 99 | 79 | 20 |

Table A.3: Number of examples (including single- and multi-findings) for each pairwise comparison for which both coders provided a code set containing Clinical Finding codes. Human coders A and B; Comp = software; GS = gold standard dataset.

## Human to human agreement (larger dataset)

To make the pairwise comparisons comparable and consistent, all the results reported in the main text refer to the Gold Standard dataset (see below). This included the comparison between two human coders (Table 1), which was performed on the examples from the Gold Standard dataset. Since we had a larger dataset coded by both coders (see Table A.2), we repeated the analysis on the subset of 291 double annotated examples. The results (Table A.4) are consistent with the results obtained on the smaller dataset (Gold Standard).

| | Number of instances | Exact match (%) | Distance (%) | | |
|---|---|---|---|---|---|
| | | | <=1 | <=2 | <=3 |
| A vs B (GS dataset) | 97 | 73 | 81 | 90 | 95 |
| A vs B (larger dataset) | 227 | 73 | 80 | 89 | 93 |

Table A.4: Comparison between human coders (A and B) - all results.
We note that the comparison on the GS dataset was on 97 instances (out of the total of 130 in the GS dataset), whereas the larger analysis was on 227 instances (out of the total 291 in the dataset). See the previous paragraph for clarifications.

## Human to computer agreement

We also performed comparisons between the human coders and the software on both the Gold Standard and entire available ("larger") datasets. We note these were direct pairwise comparisons, rather than the comparisons reported in Tables 2 and 3, which referred to the evaluation against the Gold Standard. The aim here was to understand if the human and automated coders would agree on specific instances, rather than whether the coding is correct.

The results on both the Gold Standard dataset (Table A.5) and the larger dataset (Table A.6) indicate that the average proportion of exact matches between the human clinicians and the software was around 62%, while for 88-90% of the code sets the average distance between the human and software annotated codes was 3 or less. This is almost exactly the same as the agreement between the Gold Standard codes and the software (Table 3). As expected, the agreement is notably better in the single-finding cases, where the exact matches between the codes provided by human coders and software were recorded in 75% of cases, and the distance of 3 or less in on average 96% of cases.

| | | Exact match (%) | Distance (%) | | |
|---|---|---|---|---|---|
| | | | <=1 | <=2 | <=3 |
| All | A vs Comp | 64 | 79 | 85 | 90 |
| | B vs Comp | 61 | 74 | 84 | 91 |
| | Avg A/B | 62 | 77 | 85 | 90 |
| Single-finding | A vs Comp | 75 | 93 | 95 | 95 |
| | B vs Comp | 74 | 90 | 96 | 96 |
| | Avg A/B | 75 | 91 | 96 | 96 |
| Multi-finding | A vs Comp | 6 | 13 | 38 | 63 |
| | B vs Comp | 10 | 15 | 35 | 70 |
| | Avg A/B | 8 | 14 | 36 | 67 |

Table A.5: Pairwise human to computer comparisons on the Gold Standard dataset. The averages (e.g. Avg A/B) were obtained by micro-averaging.

| | Number of instances | Exact match (%) | Distance (%) | | |
|---|---|---|---|---|---|
| | | | <=1 | <=2 | <=3 |
| A vs Comp | 347 | 62 | 73 | 82 | 88 |
| B vs Comp | 356 | 62 | 71 | 82 | 88 |
| Avg A/B | | 62 | 72 | 82 | 88 |

Table A.6: Pairwise human to computer comparisons on the larger datasets. The averages (e.g. Avg A/B) were obtained by micro-averaging.

## A.4 Coding similarity

While Section A.3 looked at the agreement between the coders even when the coding is not correct, we have further examined the level of agreement (or similarity) between the human coders and software agree with regards to the quality of coding examples. We focused only on the Gold Standard dataset and, for each free-text example, we first compared the distance between the provided codes and the Gold Standard. This has resulted in three similarity matrices (between coders A and B; coder A and software; coder B and software – see Table A.7).

|  |  | Coder B | | | | |
|---|---|---|---|---|---|---|
|  |  | Exact | 0 < D(X, Y) <= 1 | 1 < D(X, Y) <= 2 | 2 < D(X, Y) <= 3 | D(X, Y) > 3 |
| Coder A | Exact | 65 | 3 | 2 | 1 | 1 |
|  | 0 < D(X, Y) <= 1 | 3 | 3 | 2 | 0 | 0 |
|  | 1 < D(X, Y) <= 2 | 4 | 1 | 2 | 0 | 0 |
|  | 2 < D(X, Y) <= 3 | 3 | 0 | 0 | 0 | 0 |
|  | D(X, Y) > 3 | 4 | 0 | 0 | 0 | 3 |

|  |  | Comp | | | | |
|---|---|---|---|---|---|---|
|  |  | Exact | 0 < D(X, Y) <= 1 | 1 < D(X, Y) <= 2 | 2 < D(X, Y) <= 3 | D(X, Y) > 3 |
| Coder A | Exact | 55 | 8 | 2 | 2 | 4 |
|  | 0 < D(X, Y) <= 1 | 4 | 1 | 2 | 1 | 0 |
|  | 1 < D(X, Y) <= 2 | 0 | 3 | 2 | 1 | 1 |
|  | 2 < D(X, Y) <= 3 | 0 | 2 | 0 | 1 | 0 |
|  | D(X, Y) > 3 | 0 | 0 | 0 | 0 | 7 |

|  |  | Comp | | | | |
|---|---|---|---|---|---|---|
|  |  | Exact | 0 < D(X, Y) <= 1 | 1 < D(X, Y) <= 2 | 2 < D(X, Y) <= 3 | D(X, Y) > 3 |
| Coder B | Exact | 56 | 8 | 5 | 3 | 7 |
|  | 0 < D(X, Y) <= 1 | 3 | 1 | 1 | 2 | 0 |
|  | 1 < D(X, Y) <= 2 | 0 | 4 | 1 | 0 | 1 |
|  | 2 < D(X, Y) <= 3 | 0 | 1 | 0 | 0 | 0 |
|  | D(X, Y) > 3 | 0 | 0 | 0 | 0 | 4 |

Table A.7: Similarity matrices with distances D(X,Y) to the Gold Standard.
The tables show the numbers of cases rather than percentages.

The numbers on the diagonals show the similarity of the resultant codes (i.e. agreement). While in case of human coders the majority of cases are indeed on the diagonal (which can be interpreted as that there is agreement between the coders on which examples are "easy" (Exact) or "difficult" (distances over 3 from the Gold Standard)), there are several cases (the Exact column and the Exact row) of disagreement, where one of the coders have provided the correct Gold Standard code, whereas the other provided codes that are even more than 3 edges away (5 cases in total). In three instances, both human coders provided a code that was more than 3 steps away from the Gold Standard.

In the case of the agreement between human coders and the software, in almost 60% of the examples (55 out of 97) for which a human coder provided the exact match to the gold standard, the software also provided the exact match. On the other hand, for almost all cases where the software provided an exact match to the gold standard, so did the human coders, indicating that the cases that software found "easy" were also "easy" for the human coders. Similarly to the cases between human coders, there were cases (4 and 7 respectively for coder A and B) where the software provided a code that was more than three edges away (D(X, Y) > 3) from the gold standard, despite the fact that the human coder provided the exact match to the gold standard.

We have also compared the agreement of qualitative labels assigned to each free-text diagnosis, compared to the Gold Standard. Table A.8 gives the similarity matrices similar to those presented in Table A.7, but with qualitative labels. As before, the values on the diagonals represent agreements.

|  |  | Coder B | | |
|---|---|---|---|---|
|  |  | Good | Acceptable | Not acceptable |
| Coder A | Good | 80 | 3 | 2 |
|  | Acceptable | 7 | 7 | 1 |
|  | Not acceptable | 2 | 0 | 0 |

|  |  | Comp | | |
|---|---|---|---|---|
|  |  | Good | Acceptable | Not acceptable |
| Coder A | Good | 70 | 8 | 7 |
|  | Acceptable | 5 | 6 | 4 |
|  | Not acceptable | 1 | 0 | 1 |

|  |  | Comp | | |
|---|---|---|---|---|
|  |  | Good | Acceptable | Not acceptable |
| Coder B | Good | 70 | 10 | 9 |
|  | Acceptable | 6 | 3 | 1 |
|  | Not acceptable | 0 | 1 | 2 |

Table A.8: Similarity matrices with qualitative labels as compared to the Gold Standard. The tables show the numbers of cases rather than percentages.

The codes assigned by the human coders agree with the Gold Standard in most instances (85% (87/102) of cases). Still, there are few cases (4 in total) where one of the coders provided a Good code whereas the other coder provided a Not acceptable code for the same textual description. When compared to the software, there is a larger discrepancy between the labels assigned to codes from the human and software coders. Still, on average in 75% (76 / 102) cases, the codings provided by both the human coder and the software were of the same quality according to the panel's qualitative assessment.  In 8 cases on average (7 and 9 for coders A and B respectively), the code provided by the human coder was considered Good but the software struggled to capture clinical intent (Not acceptable). Conversely, there was a single example for which a code assigned by a human coder received a rating of "Not acceptable", while the software coding was rated as "Good". In 1-2 cases, both the human and software coders provided Not acceptable codes for a given textual description.

## A.5 Coding guidelines

Guidelines for manual SNOMED CT coding of free-text diagnoses

### 1) Task Overview

The coding task involves the assignment of one or more SNOMED CT identifiers (SCTIDs) to a given free-text diagnosis. We will code diagnoses that have been noted/listed in a clinical letter under a *Diagnoses* heading, rather than coding diseases in real-time settings (e.g. during a consultation). Thus, the coding task involves some interpretation of the clinical intent expressed by a free-text diagnosis expression. As a coder, you will use your clinical judgement to find the code(s) for the most suitable concept(s) that reflect the likely clinical intent in a particular diagnosis.

We concentrate only clinical findings i.e. disorders (including problems, diseases). All assigned (core) clinical concepts should therefore be of type *Disease (disorder)* (in addition to any qualifiers). Note that in this exercise we will <u>not</u> code other SNOMED CT concept types, e.g. *procedures*, *situations, social context* etc.

### 2) General coding strategies – what and how to code

### A. Code disorders specified in a free-text diagnosis

The main task is to identify and code clinically relevant disorders/diseases/problems that are explicitly mentioned in a free-text diagnosis. While we aim to code likely clinical intent, we do not want to infer (any additional, not explicitly mentioned) disorders from free-text expressions. Still, we will aim to code as much of the context as possible, in particular if there is an existing, pre-coordinated SNOMED CT concept that corresponds to the stated disorder.

If a free-text diagnosis explicitly mentions a clinical procedure, we can code it but it will not be used in the analyses. Again, do not infer a problem based on the nature of a stated procedure, but if the problem is explicitly stated in the wording of the procedure, then code the problem. For example, in "*cataract surgery*", we will code "*cataract*" as a problem, and optionally *cataract surgery*" as procedure. However, don't make assumptions: *CABG*" would only be coded as a procedure and we should not infer that there is underlying CAD as a problem; similarly, "*appendectomy*" would only be coded as a procedure and not also as "appendicitis".

> **Note 1:**
> Please also note that in metonymic cases such as, for example, when the name of a virus is used to describe the associated disorder, it is important to refine the search query in order to retrieve the SCT code for the associated disorder rather than just record the code for the virus (which will probably be the code of an organism).
>
> **Example: 'E. Coli' should not be coded with a code for an organism.**

## B. Pre-coordination

For many health conditions and procedures, there are pre-defined SNOMED CT concepts, which fully present quite specific health states, diagnoses or interventions. For example, there is a single code for "*Seropositive errosive rheumatoid arthritis*" (SCTID: 308143008). In this case a **single, predefined** SNOMED CT concept that can be used to describe the meaning of the diagnosis. This coding strategy is called **pre-coordination**, and it is the **preferred** way to code diagnoses: *whenever possible, select a single concept to code a given diagnosis or procedure*.

**Example:** "*sever asthma*" should be mapped to *Severe asthma (disorder) SCTID: 370221004,* rather than mapping it to two separate concepts: "*severe*" SCTID 24484000 and "Asthma (disorder)" SCTID: 195967001.

> **Note 2:** Although pre-coordination and post-coordination (see below) can result in equivalent semantic descriptions for the same diagnosis, underline{pre-coordination is the preferred coding option} for this task and it should be explored before post-coordination. The pre-coordinated terms should be of type *disorder* as only these codes will be used for analyses.

## C. Post-coordination

There might be cases where there is not a single concept that describes a given diagnosis. In that case, we can **combine** two or more concepts. This strategy is called **post-coordination**. Note that one of these codes has to be of the disorder type (which will be used for analyses), but other codes can be used for clarity and/or completeness.

**Example:** There is not a single code to represent "*severe headache*"; instead, we need to use two codes, one to represent "headache" (SCTID 25064002) and a separate code for the modifier "*severe*" (SCTID 24484000).  These two codes are then placed together as post-coordination combination of codes 24484000 and 25064002.

When coding a diagnosis for which there is not a single pre-coordinated concept that captures all the aspect of a given diagnosis, we should identify *modifiers* to the core concept, including:

- Locus/finding site e.g. "liver disease"

- Laterality e.g. "*right* eye infection"
- Severity e.g. "*severe headache*"
- Chronicity/temporal associations e.g. "post-viral disorder"
- Finding method e.g. "lung cancer detected by biopsy"
- Causative associations e.g. "pancreatitis due to infection"

and then aim to establish whether there is a concept that captures at least some of these modifiers together with the core concept in a pre-coordinated concept. It is suggested that the modifiers are checked in the order specified above (i.e. give priority to body structure/laterality, then chronicity and then severity). A good strategy is to look at the hierarchy starting from the core disease term and see if any additional modifier has been already pre-coordinated.

**Example:** *"Mild right sacroiliitis"* should be coded by post-coordinating

- Inflammation of sacroiliac joint (disorder); SCTID: 55146009
- Right (qualifier value); SCTID: 24028007
- Mild (qualifier value); SCTID: 255604002

Note that we prefer using "Right (qualifier value)" instead of "Structure of right sacroiliac joint (body structure); SCTID: 722778007" to avoid repeating information (given that "Inflammation of sacroiliac joint" already include the information about the body part).

**Example:** *"fractured right arm"* is made up of the core concept "*arm fracture*" and laterality qualifier "*right*":

- Fracture of upper limb (disorder); SCTID: 23406007
- Right (qualifier value); SCTID: 24028007

So, the coding here is a post-coordination of two SCTIDs: 23406007+24028007.
An alternative (but not preferred) coding is to consider the core concept "*fracture*" and laterality indicator "*right arm*":

- o Fracture of bone (disorder) SCTID: 125605004
- o Right upper arm structure (body structure) SCTID: 368209003

However, note that "Right upper arm structure (body structure)" may refer to "upper arm", which might be misleading.

**Example:** "*Cataract surgery (right eye)*" should be coded as
- Cataract (disorder); SCTID: 193570009
- Right (qualifier value); SCTID: 24028007

Note also that qualifier *right* cannot be combined with a procedure, so if the procedure is coded, then it should be coded as:

- Cataract surgery (procedure); SCTID: 110473004
- Right eye structure (body structure); SCTID: 18944008

rather than

- Cataract surgery (procedure); SCTID: 110473004
- Right (qualifier value); SCTID: 24028007 [procedure can't be right or left]

**Example:** Temporal context should be encoded using a relevant qualifier; for example, "*Previous pulmonary embolism*" should be coded as

- Pulmonary embolism (disorder); SCTID: 59282003
- Previous (qualifier value); SCTID: 9130008

**Example:** Suspected diagnoses should be coded using "Probable diagnosis (contextual qualifier) (qualifier value); SCTID: 2931005" if there is not an appropriate pre-coordinated term. For example, **"***Likely primary Raynaud's***"** should be coded as

- Isolated primary Raynaud's phenomenon (disorder); SCTID: 361131008
- Probable diagnosis (contextual qualifier) (qualifier value); SCTID: 2931005

> **Note 3:** As a rule, in the case of a post-coordinated expression, find first the SCTID of the core disease/disorder concept, followed by the SCITDs of concepts that are used to supplement, refine or modify the meaning of the core disease concept. Use a combination that has a minimal number of concepts and avoids duplication of information.

## D. Distinct clinical concepts

When two or more **distinct** clinical concepts are present in the same narrative description, these should be coded as separate concept. Special care should be taken not to confuse the situation with that of post-coordination. A typical example is a disorder and an associated procedure, which should be coded separately as two annotation concepts.

**Example:** "*Anxiety and Depression*" make up two different concepts that should be mapped to separate SNOMED concepts:

- Anxiety disorder (disorder); SCTID: 197480006
- Depressive disorder (disorder) SCTID: 35489007

These two concepts are **not** post-coordinated: there are two separate codes for two diagnoses. Clinical judgement should be used to establish that a given description is about two (or more) conditions, rather than one.

In cases where there is a pre-coordinated term that combines disorders (see below for examples), we will prefer the pre-coordinated term.

**Example: "***Prior MI and stents***"** **should be coded as**

- Myocardial infarction (disorder); SCTID: 22298006
- Prior diagnosis (contextual qualifier) (qualifier value); SCTID: 48318009

and additionally *'stents'* as a procedure:

- Insertion of arterial stent (procedure); SCTID: 233404000

**Example:** "*Pancreatitis due to gallstone*" should be coded as a <u>single</u> pre-coordinated diagnosis ("Gallstone pancreatitis (disorder); SCTID: 95563007"), rather than "Pancreatitis (disorder); SCTID: 75694006" and "Gallbladder calculus (disorder); SCTID: 235919008". Similarly, *"CKD stage 1 due to hypertension"* should be coded as one concept: "Chronic kidney disease stage 1 due to hypertension (disorder); SCTID: 117681000119102".

---

**Note 4:** The rules for pre/post-coordination would still apply for each separate concept.

**Note 5**: In cases when the same concept is repeated in the same clinical description (for example, when a clarification is given within a parenthetical expression), just record the associated SCT code only once.

**Note 6:** In cases of multiple concepts which are synonymous to each other just record the code for one of them (preference should be given to disorder type and then to the most specific concept in the hierarchy). Do not use OR or AND Boolean operators to signify synonymy or conjuction of concepts.

---

**E. Parent vs. child**

If there is uncertainty about choosing between a more general (i.e. parent concept) and a more specific concept (i.e. child), go for the more specific concept if applicable. If candidate SNOMED CT concepts are children of the same parent, use the SCTID of the parent concept. For example, if there is uncertainty about choosing among 'rheumatic arteritis' and 'senile arteritis', then use the SCTID of the parent concept  'arteritis'.

**3) General guidelines – what and how <u>*not*</u> to code**

**F. Do <u>not</u> code diagnoses as <u>situations</u>**

SNOMED CT provides *Situations* as a type of (pre-coordinated) concept that specifically includes a definition of the context of use of a clinical finding or procedure. We will not use situations for coding; rather whenever we have a disorder (procedure), code it as a *Disorder* (or *Procedure*) and post-coordinate if necessary with relevant qualifiers. The main reason for this is practical: the aim of our exercise is to evaluate coding of diagnoses (and not situations).

**Example:** History or past diagnosis should be coded using a relevant disorder term and a suitable qualifier, even when there is a pre-coordinated <u>situation</u>. For example, "*History of hypertension*" should be coded as

- Hypertensive disorder, systemic arterial (**disorder**); SCTID: 38341003
- History of (contextual qualifier) (qualifier value); SCTID: 39252100

rather than "History of hypertension (**situation**); SCTID: 161501007". Similarly, "Past hypertension" should be coded as

- Hypertensive disorder, systemic arterial (disorder); SCTID: 38341003
- In the past (qualifier value); SCTID: 410513005

Note that there isn't a pre-coordinated term here. Also, use lexically closest qualifier to the one that appeared in the free-text diagnosis as long as it satisfies the clinical intent (so, if '*past*' appears in the description, use it rather than '*history of*' to find a suitable qualifier).

### G. Do <u>not</u> post-coordinate specific finding/diagnostic methods with diagnoses

If specific finding methods and measurements are mentioned as part of diagnosis descriptions, do not post-coordinate them with the main diagnoses. However, if there is a pre-coordinated concept that captures the whole description, use it as more appropriate.

**Example:** "*Chronic renal impairment (eGFR 44)*" should be coded as

*Chronic kidney disease (disorder) SCTID: 709044004*

Note that there is a code for "*eGFR (observable entity)* SCTID: 80274001"; while we could post-coordinate the diagnosis (SCTID: 709044004) with the finding method (eGFR is a Finding Method i.e. a permissible attribute for a clinical finding), we will not code that in this exercise. However, "*Chronic renal impairment (stage 1)*" should be coded as "Chronic kidney disease stage 1 (disorder); SCTID: 431855005"

Similar examples (code only the underlined parts):
<u>Osteopenia</u> on DEXA – 2012
<u>Osteoporosis</u> – DEXA 2014 T score -3.2 spine, -3.7 femur
<u>Mild right sacroiliitis</u> (from previous MR scan)

Still, use judgement to code any finding/diagnostics that has clinical significance (e.g. of a particular diagnostic value would indicate a necessary severity modifier).

**H. Do not post-coordinate drugs/treatments associated with a disease**

In this task we will not code drugs or treatments unless they have been pre-coordinated as a disorder term.

**Example:** "*Atrial fibrillation (on Warfarin)*" we should not code "*Warfarin*", rather only

- Atrial fibrillation (disorder); SCTID: 49436004

**Example:** "*Treated vitamin D deficiency 2012*" should be coded only as

- "Vitamin D deficiency (disorder); SCTID: 34713006"

**Example:** We will ignore the context about drugs or treatments unless there is a pre-coordinated disease term. For example, "*Hypertension caused by contraceptive pill*" should be coded as

- "Hypertension caused by oral contraceptive pill (disorder); SCTID: 169465000".

**I. Do not code *explicit temporal* context (e.g. dates)**

Do not code explicit temporal information (e.g. dates of diagnoses or procedures), but code qualifiers such as *recent, prior, history of* (see examples above). In cases where a procedure is planned for future, do not encode such procedures at all.

**Example:** "*Right total hip replacement February 2015*" should be coded as

- Total replacement of right hip joint (procedure); SCTID: 443435007

  *Note*: there is a code for *total hip replacement* (Total replacement of hip (procedure); SCTID: 52734007) as well for '*right hip*' (Right hip region structure (body structure) SCTID: 287579007), but we prefer a pre-coordinated term, rather than post-coordination.

Relevant temporal context can be still coded using appropriate qualifiers (but note that these will not be used for evaluation).

**Example:** in "*Recent eye cataract surgery*", we will code "*Recent*" as a qualifier

- Cataract surgery (procedure); SCTID: 110473004
- Recent (qualifier value); SCTID: 6493001

Note that we will also code the associated (explicit) disorder:

- Cataract (disorder); SCTID: 193570009

**J. Do <u>not</u> post-coordinate causative associations** (unless pre-coordinated)

**Example**: "*pancreatitis due to infection*" should be only coded as "*pancreatitis*". However, as indicated above, "*Pancreatitis due to gallstone*" should be coded as a <u>single</u> diagnosis ("Gallstone pancreatitis (disorder); SCTID: 95563007") as there is a pre-coordinated term.

**K. Do <u>not</u> post-coordinate information that is redundant/obvious**

**Example:** in "*Likely primary Raynaud's in hands and feet*", since *Raynaud's* can only be in hands and feet, aim to code only the rest of the diagnosis ("*Likely primary Raynaud's*").

**L. Do <u>not</u> code *outcomes***

**Example:** in "*Recent right tennis elbow, improved on recent injection*" – we will not code that the problem has "*improved*". We will still however code two clinical concepts (procedure and problem), with optional temporal qualifiers (i.e. *recent* – not shown below):

- Tennis elbow injection (procedure); SCTID: 274496007
- Right elbow region structure (body structure); SCTID: 368149001
  [Note: We cannot use Right (qualifier value); SCTID: 24028007 as procedures can't be right or left.]

and

- Lateral epicondylitis (disorder); SCTID: 202855006.
- Right (qualifier value); SCTID: 24028007

**4) Summary: finding the right code(s) for a diagnosis free-text description**

**Main principles:**

1. The main task is to code free-text **diagnoses**. Capture as much as possible in a single pre-coordinated disorder term. If free-text explicitly refers to a **procedure**, code it as a separate concept if needed for completeness and clarity.
2. Explore the possibility of a pre-coordinated term for the whole description, ignoring the parts we do not want to encode.
3. Identify if a single disorder is expressed in a free-text diagnosis. If not, apply the steps to separate core diagnoses (and procedures).

**Steps:**

**Step 0:** Check if the free-text description contains any clinically relevant concept; this is to save time and avoid errors because in some case there is just a term like *'weight'* or *'history'* in the data.

**Step 1:** Check pre-coordination first; this is the preferred option and should be checked before anything else, even if there is suspicion about post-coordination in the description.

> If the above steps fail, use synonyms, abbreviations or parent terms for the core concept (see also tips below).

**Step 2:** Checking for multiple (distinct) concepts as discussed in the guidelines; the point here is to decompose the term and search for pre-coordination for each concept separately again (before moving to post-coordination for each of them if needed).

**Step 3:** Post-coordinate if needed; since post-coordination involves checking for codes of the core concepts and its qualifiers separately, you can apply Step 1 (i.e. similar to doing pre-coordination for core concept only, pre-coordination for a qualifier only etc.).

**Tips:**

- Try synonyms: It is likely that a clinical term is expressed as a synonym, abbreviation or even a 'lay term' of a SNOMED CT concept. For example, in the area of rheumatology, 'Osteonecrosis' may be used as a synonym of 'Avascular Necrosis' and 'CPDD' an abbreviation of 'Calcium Pyrophosphate Dihydrate Crystal Deposition Disease'.

- Try longer forms of the term – quite often, they are described as synonyms of an existing SNOMED CT concept and the browser might match that.

  **Example:** 'CFIDS' may return nothing, but 'Chronic Fatigue' might give matches

- If nothing is returned, try stripping plurals, modifiers (*left*, *right*, etc.), or try with just a couple of letters and then browse the hierarchy to find a good match.

- Terms that include words such as: "*and*", "*or*", "*with*" etc. may contain multiple concepts and may need to be searched separately; search each term by breaking it up into core concept (e.g., head noun) and other term components.